\documentclass[journal]{IEEEtran}
\usepackage{blindtext}
\usepackage{graphicx}
\usepackage{color}
\usepackage{gensymb}

\ifCLASSINFOpdf
\else
\fi
\begin{document}
%\title{Real-Time Human Detection as an Edge Service Enabled by a Lightweight CNN}

\title{Smart Surveillance as an Edge Network Service: from Harr-Cascade, SVM to a Lightweight CNN}


\author{Seyed Yahya Nikouei, Yu Chen, Sejun Song, Ronghua Xu, Baek-Young Choi, Timothy R. Faughnan
% <-this % stops a space
\thanks{S. Y. Nikouei, Y. Chen, and R. Xu are with the Department
of Electrical and Computer Engineering, Binghamton University, SUNY, Binghamton,
NY 13902 USA, email: \{snikoue1, ychen, rxu22\}@binghamton.edu.}% <-this % stops a space
\thanks{S. Song and B. Y. Choi are with School of Computing and Engineering, University of Missouri-Kansas City, Kansas City, MO 64110 USA, email: \{songsej, choiby\}@umkc.edu.}% <-this % stops a space
\thanks{T. R. Faughnan is with the New York State University Police, Binghamton University, SUNY, Binghamton, NY 13902 USA, email: tfaughn@binghamton.edu.}% <-this % stops a space
%\thanks{Manuscript received in March 2018.}
}

\maketitle

\begin{abstract}
Edge computing efficiently extends the realm of information technology beyond the boundary defined by cloud computing paradigm. Performing computation near the source and destination, edge computing is promising to address the challenges in many delay-sensitive applications, like real-time human surveillance. Leveraging the ubiquitously connected cameras and smart mobile devices, it enables video analytics at the edge. In recent years, many smart video surveillance approaches are proposed for object detection and tracking by using Artificial Intelligence (AI) and Machine Learning (ML) algorithms. 
%However, it is still hard to migrate those computing and data-intensive tasks from Cloud to Edge due to the high computational requirement and still Many challenges face decentralized smart surveillance application at edge. 
This work explores the feasibility of two popular human-objects detection schemes, Harr-Cascade and  HOG feature extraction and SVM classifier, at the edge and introduces a lightweight Convolutional Neural Network (L-CNN) leveraging the depthwise separable convolution for less computation, for human detection. Single Board computers (SBC) are used as edge devices for tests and algorithms are validated using real-world campus surveillance video streams and open data sets. The experimental results are promising that the final algorithm is able to track humans with a decent accuracy at a resource consumption affordable by edge devices in real-time manner.

\end{abstract}


% \fi
\IEEEpeerreviewmaketitle

\section{Introduction}

Nowadays almost every person can be connected to the network using their pocket-sized mobile devices wherever and whenever. The advancement of cyber-physical technologies and their interconnection through elastic communication networks facilitate the concept of the Smart Cities that improve the life quality of residents. Attracted by the convenient lifestyle in bigger cities, the world's population has been increasingly concentrated in urban areas at an unprecedented scale and speed~\cite{chen2016smart}.

The fast pace of urbanization \cite{chen2016dynamic} poses many opportunities and challenges. The recent concept of Smart Cities has attracted the attention of the urban planners and researchers to enhance the security and well-being of the residents. The proliferation of information and communication technologies (ICT) connects cyber-physical systems and social entities as well as facilitates many smart community services. One of the most essential smart community services is the intelligent resident surveillance \cite{cenedese2014padova}. It enables a broad spectrum of promising applications, including access control in areas of interest, human identity or behavior recognition, detection of anomalous behaviors, interactive surveillance using multiple cameras and crowd flux statistics and congestion analysis and so on \cite{hu2004survey}. 

Many of these smart surveillance applications require significant computing and storage resources handling massive contextual data created by video sensors. A typical low frame rate (1.25 Hz) wide area motion imagery (WAMI) sequence alone can generate over 100M of data per second (400G per hour). According to the recent study, the video data dominates the real-time traffic and creates heavy workload on the communication networks. For example, online video accounts for 74\% of all online traffic in 2017 and 78\% of mobile traffic will be video data by 2021 \cite{cisco2017}. Thus, it is important to handle this massive data transfer in new ways. The cloud computing paradigm provides excellent flexibility and is also scalable corresponding to the increasing number of surveillance cameras. In practice, however, there are significant hurdles for the remote cloud-based smart surveillance architecture. 

Key surveillance applications such as monitoring and tracking need a real-time capability. However, processing raw video data from widely distributed video sensors such as Close-Circle Television (CCTV) cameras and mobile cameras not only incurs uncertainty in data transfer and timing but also poses significant overhead and delay to the communication networks\cite{chen2016smart}. Also, it may cause the data security and privacy issues by providing more attacking opportunities for adversaries. Therefore, current surveillance applications are for off-line forensics analysis instead of a proactive tool to deter suspicious activities before the damages are caused. 

The surveillance community has been aware of the growing demand for human resources to interpret the data due to the ubiquitous deployment of networked static and mobile cameras and has made many efforts in past decades~\cite{ma2017survey}. For example, many automated anomaly detection algorithms have been investigated using machine learning~\cite{ribeiro2017study} and statistical analysis~\cite{fuse2017statistical} approaches. Although these intelligent approaches are powerful, they are computationally very expensive. Hence they are implemented as a central cloud service. Researchers are also trying to help operation personnel beware events using event-driven visualization mechanism~\cite{fan2017heterogeneous}, re-configuring the networked cameras~\cite{piciarelli2016dynamic}, and mapping conventional real-time images to 3D camera images~\cite{jiewu1}. However, the traditional human-in-the-loop solutions are still challenged significantly by the demand of real-time surveillance systems due to the lack of scalability. For example, video analysis mostly relies on teams of specially trained officers manually watching thousands of hours of different format and quality videos, looking for one specific target. Due to the manual coordination and tracking and mechanical pan-tilt-zoom (PTZ) remote control, it is challenging to achieve adequate real-time surveillance.

Edge computing as a surveillance service is considered as the answer to the shortcomings \cite{ahmed2017mobile}, \cite{ali2017sedasc}, \cite{cao2017edge}. The edge computing technology migrates more computing tasks to the connected smart “things” (sensors and actuators) at the edge of the network \cite{shi2016edge}. In general, edge computing possesses the following advantages compared to cloud computing:

\begin{enumerate}
\item \textit{Real-time response}: applications or services are directly executed on-site or near-site, communication delays are minimized, which is essential to delay sensitive, mission critical tasks, such as the smart surveillance; 
\item \textit{Lower network workload}: raw data generated by sensors or monitors is consumed at the edge of the network instead of outsourcing to a remote cloud center. While the processed results may be sent to the cloud for future analysis, the communication overhead is much lower than outsourcing tasks to cloud; 
\item \textit{Lower energy consumption}: most of the edge devices are energy constrained, by its nature the algorithms deployed at the edge are lightweight that will reduce energy consumption for the process and data transmission in total; and
\item \textit{Data security and privacy}: the less data is sent, the fewer opportunities are available to adversaries to compromise the confidentiality and integrity of the data, also it is easier to enforce security and privacy policies at local network in comparison to requesting collaboration among multiple network domains under different administrations. 
\end{enumerate}

In this paper, we propose to devolve more intelligence to the edge to significantly improve many smart tasks such as fast object detection and tracking. Adopting the recent research on machine learning, we choose the Convolutional Neural Network (CNN) algorithm which incurs comparatively less pre-processing overhead than other human image classification algorithms. We efficiently tailored the CNN to be furnished in the resource-constrained edge devices according to an observation that surveillance systems are mainly for the safety and security of human being. According to our experimental study, the lightweight CNN (L-CNN) algorithm can process an average of 1.79 and up to 2.06 frames per second (FPS) on the selected edge device, a Raspberry PI 3 board. It meets the design goals considering the limited computing power and the requirement from the application.

In summary, the major contributions of this work are highlighted below:

\begin{enumerate}
    \item Aiming at intelligent surveillance as an edge network service, a thorough study of two well-known human-objects detection schemes, Harr-Cascade and HOG+SVM, has been conducted, which evaluates their feasibility of running on resource-limited edge devices;
    
    \item A lightweight Convolutional Neural Network (L-CNN) is applied to enable a real-time human-objects identification on a network edge \cite{nikouei2018real}; 
    
    \item Instead of simulation, system-oriented research has been conducted. The L-CNN, SSD GoogleNet, Harr-Cascade, and HOG+SVM algorithms are implemented on a Raspberry PI Model 3 board as the edge device; and  
    
   \item An extensive experimental validation study has been conducted using real-world surveillance video data. Comparing with SSD GoogleNet, Harr-Cascade and HOG+SVM, the L-CNN is a promising approach for delay-sensitive, mission-critical applications like real-time smart surveillance.  
    
\end{enumerate}

\begin{figure}[t]
    \centering
        \includegraphics[width=0.425\textwidth]{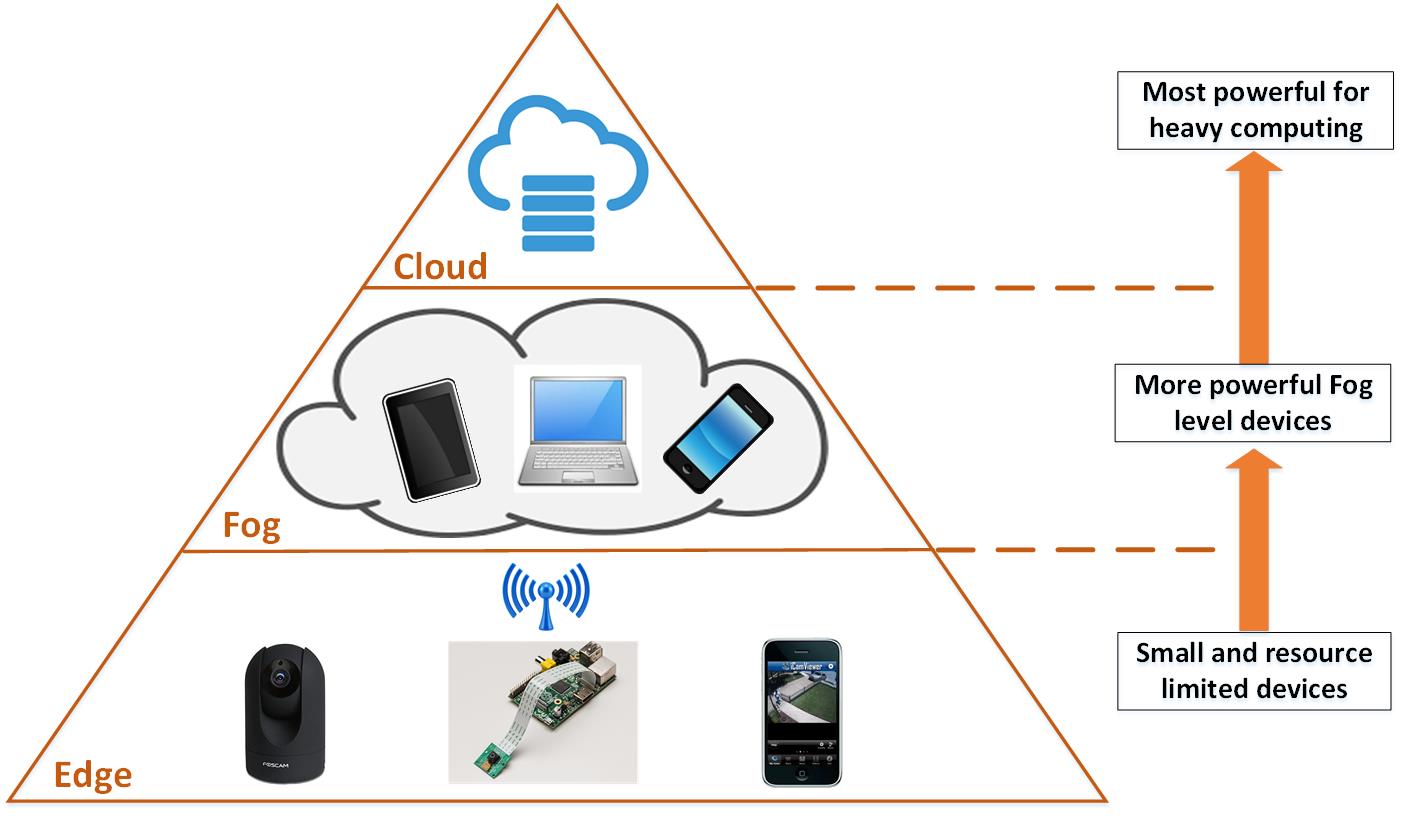}
    \caption{Edge-Fog-Cloud hierarchical architecture.}
    \label{fig:arch}
    \vspace{-15pt}
\end{figure}

The rest of the paper is sorted as follows. Section \ref{sec:background} provides background of the closely related work. Section \ref{sec:edge} explains Haar-Cascaded and HOG+SVM at edge. Then, Section \ref{sec:L-CNN} introduces the proposed lightweight CNN architecture, and the training of the L-CNN is disused in Section \ref{sec:training}. Section \ref{sec:exp} explains the results of the tracking algorithm implemented on a Raspberry PI 3 model B and a Tinker board. At last, Section \ref{sec:conclusions} wraps up this paper with conclusions and discussions of our on-going efforts. 

\section{Background Knowledge and Related Work}
\label{sec:background}

\subsection{Smart Surveillance as an Edge Service}

The surveillance community has been aware of the growing demand for human resources to interpret the data due to the ubiquitous deployment of networked static and mobile cameras and has made many efforts in past decades~\cite{ma2017survey}.

Traditional surveillance systems depend on human operators to manipulate the processing of captured video \cite{chamasemani2013systematic}. However, there are many shortcomings with this approach. Not only it is unrealistic to have a human operator to maintain full concentration on the video for a long time, but it is also not scalable as the number of cameras as sensors grows significantly. More recently proposed smart systems considered as the second generation of the surveillance systems aimed at minimizing the role that human operators play in object detection, and the responsibility of abnormal behavior detection is taken by various more intelligent machine learning algorithms~\cite{chen2016dynamic},~\cite{wang2013intelligent}. The algorithm automatically processes the collected video frames in a cloud to detect, track, and report any unusual circumstances.

Figure \ref{fig:arch} presents an edge-fog-cloud hierarchical architecture in which functions in a smart surveillance system are classified into three levels: 

\begin{itemize}
    \item Level 1: each object of interest is identified through low-level feature extraction from video data;
    \item Level 2: the behavior or intention of each object of interest is detected/recognized, quick alarm raising; and
    \item Level 3: anomalous or suspicious activities profile building and historical statistical analysis and also fine tuning through online training the decision making algorithm.
\end{itemize}

Ideally, the minimum delay and communication overhead would be achieved if all the functions are conducted on-site at the network edge where the sensor is located, and the decision is made instantly~\cite{chen2017enabling},~\cite{chen2016smart}. However, it is not realistic to accomplish the operations of Level 1 and Level 2 by the edge devices. Therefore, once the detection and tracking tasks are done, the results are outsourced to the fog layer for further data contextualization and decision making. The computationally expensive Level 3 functions can be positioned on the fog computing level or even further away on the cloud centers considering the constraints on edge processing power. And functions like long term profile building based on geo-location of the camera is not required to be accomplished instantly. In a smart surveillance system, the Level 1 functions are the fundamental. More specifically human object detection is vital as missing any objects in the frame will lead to undetected behavior. Also. the false positive rate should be minimized, because wrongly identifying an object in the frame as a human will possibly result in false alarms.

\subsection{Human-Object Detection}

Haar-like feature extraction is well suited for face and eye detection~\cite{cristani2013human}. Haar models are light weighted and very fast, which are appreciated as a candidate for edge implementation. However, the human body can have different appearance in different ambient lighting, which is harder for this type of models to achieve a high detection accuracy~\cite{hannuna2016ds}. 

Grids of Histograms of Oriented Gradient (HOG) can produce reliable features for human detection~\cite{dalal2005histograms}. While the Haar features fail to detect humans when the body angle toward the camera changes, HOG features continue to perform well. HOG features are given to a Support Vector Machine (SVM) classifier to create a human detection algorithm called HOG+SVM~\cite{lowe2004distinctive}. One downside to using HOG feature extraction at the edge is that this method creates a burden on the limited resource environment. Performance results are discussed and compared in Section~\ref{sec:exp}. 

Scale Invariance Feature Transformation (SIFT) is another well-known algorithm for human detection through extracting distinctive invariant features from images, which provides features that can be used to perform reliable matching between different views of an object or scene~\cite{henriques2015high}.

\subsection{Machine Learning at the Edge}

Powerful machine learning algorithms are recognized as the solution to take full advantage of big data in many areas~\cite{kolomvatsos2017reinforcement},~\cite{zhou2017machine}. However, when the big data comes to the edge, the demand for computing and storage resources makes them unfit. The edge environment necessitates lightweight but robust algorithms. Applications of some simple machine learning algorithms have been investigated in environments with constraints on resources, such as wireless sensor networks (WSNs) and IoT devices~\cite{ahn2016machine}. There are efforts to build large-scale distributed machine learning systems to leverage heterogeneous computing devices and reduce the communication overhead~\cite{abadi2016tensorflow},~\cite{sun2016timed}. Although none of the reported algorithms is ideally fit to the resource-constrained edge environment, they have laid a solid foundation for us. In fact, the community has recognized the importance of efficient models for mobile and embedded applications~\cite{anisimov2017towards},~\cite{howard2017mobilenets},~\cite{zhang2017shufflenet}.   

GoogleNet~\cite{szegedy2015going} and Microsoft ResNet~\cite{he2016deep} are widely used and well-known architectures for image classification because of their high accuracy. They can take a picture as an input and conduct classification for up to one thousand different objects. This type of network has as many filters as needed to create a feature map that can differentiate between the possible objects classes~\cite{hinton2006reducing},~\cite{hu2015deep}. If only human objects are required to be classified or detected, the network architecture needs fewer filters to reach the same performance accuracy. However, the authors could not find any deep learning network specially designed for detecting human objects in mind, rather the networks tend to have a more generalized use cases. 

Recent attempts have been made to generate faster deep learning networks that require less resource without losing performance. As its name implies, SqueezeNet achieves the same performance as the AlexNet but takes less memory~\cite{iandola2016squeezenet}. 
MobileNet is another architecture created to work on resource constraint devices~\cite{howard2017mobilenets}. It is not only memory efficient, but also it runs very fast because of a different convolutional architecture that creates it. It has been mathematically proven that this network creates less computational burden while having fewer parameters ~\cite{sifrerigid}. MobileNet produces results comparable to GoogleNet which is of the best performing architectures in terms of accuracy. 

\section{Harr-Cascade and SVM at the Edge}
\label{sec:edge}
In this paper, we focus on fast and accurate detection of humans as objects of interest (human objects) as it is vital for the algorithm to give out the exact position coordination of the object of interest for tracking purposes. Otherwise, an abnormality detection algorithm based on human behavior may not function properly in case of incomplete information provided to it. Although discussed comprehensively in literature, in this section an overview of the Harr-Cascaded and HOG+SVM algorithms are provided. Their wide usage for human detection in surveillance, makes them noticeable candidates for edge application and exploration gives insight about their weaknesses on the edge devices.

\subsection{Harr-Cascade}
\label{sec:harr}

\begin{figure}[t]
    \centering
        \includegraphics[width=0.3\textwidth]{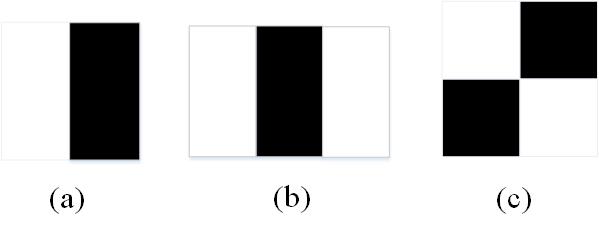}
    \caption{Examples of Haar-like features. (a) two rectangular features. (b) three rectangular features. (c) four rectangular features}
    \label{fig:Haar}
    \vspace{-15pt}
\end{figure}

Haar-like features consist of three general shapes. Figure \ref{fig:Haar} shows Haar-like feature set examples. These filters are going to convolute over an input image and in each position the sum of pixel values in black rectangles will be subtracted from the sum of pixel values in white rectangles. When the capturing angle changes the same filter may produce very different results. A simpler classifier may miss an object if the features are not totally reliable for detection. One may also argue that with a better image set for training, Haar Cascade will provide more accurate results \cite{lienhart2002extended}. 

When all possible scenarios are considered, even a $24 \times 24$ image will produce more than 160 thousand features since filters in Fig. \ref{fig:Haar} can have any combination of sizes and rotations and positions. The learning process is computationally expensive and needs to take place at CPU clusters, which might not be available to many. However, once the training is finished, a feature set is ready and only the selected features are stored for future classification. Thus, the computational complexity of the overall algorithm is small in executtion phase. 

In trainig phase best performing features are selected. The process of selecting best features is performed by the Adaboost algorithm which stands for Adaptive Boosting and it is constructed from classifiers that are called "weak learners". This algorithm generates a weighted sum between results of weak learners  (Eq. \ref{eq:adaboost}), where $h(x)$ is considered as each weak learner for input $x$. During the learning process each weak learner receives a weight in summation for error calculation (Eq. \ref{eq:adaboost_learn}), which is based on lastly calculated boosted classifier. The goal is set to minimize error value (as shown in \ref{eq:adaboost_learn}) where $i$ is every input for learning iteration $t$. 
 
\begin{equation}
F_T(x) = \sum_{t} f_t(x)  , where   f_t(x) = \alpha_th(x)
 \label{eq:adaboost}
\end{equation}

\begin{equation}
E_e = \sum_{i}E [ F_{t-1} +\alpha_th(x_i)]
 \label{eq:adaboost_learn}
\end{equation}

Positive and negative images, are collected for training, where positive images contain the object of interest with different backgrounds including the positions and coordinates of the sample. In practice, many images are used more than one time by mirroring them or cutting its edges. Negative images do not include the object of interest. In the training around 2000 positive and 1000 negative images are used. The result of training creates a file containing the specific best performing features to be executed on input images. According to the results, regid regression is further made, in area where features give positive results, to give more accurate coordination as the output.

\begin{figure}[b]
    \centering
        \includegraphics[width=0.35\textwidth]{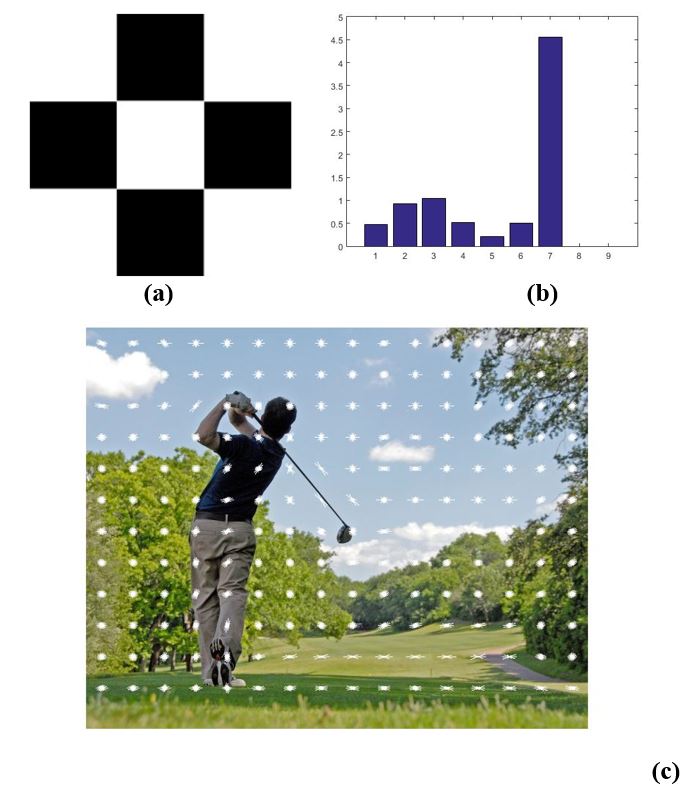}
    \caption{(a) HOG convolution filter. (b) HOG representation of an Image. (c) cell representation of the HOG~\cite{liu2016ssd}.}
    \label{fig:HOG}
    \vspace{-15pt}
\end{figure}

As revealed by the training procedure and simplistic working flow of the algorithm, Haar-Cascade object classification will not perform well if the training set does not contain all possible angles as shown in section \ref{sec:exp}. Also if the object of interest is far away from the camera, which is the usual case in surveillance application videos for outdoor applications, the algorithm may fail to detect the object. Furthermore, the simple structure may imply the loss of robustness. However it is used for low power and real-time applications because of its fast detection.

\subsection{SVM Classifiers}
\label{sec:SVM} 

Pixel values cannot be trusted because of so many parameters that affect them, other features are thus searched for.Figure \ref{fig:HOG}(a) depicts a filter that is placed on each pixel in white with its four neighboring pixels in black. X and Y derivatives are calculated simply by subtracting the horizontal neighboring and vertical neighboring pixel values respectively corresponding to the white pixel. In particular, the X derivatives are fired by the vertical lines, and Y derivatives are fired by horizontal lines, which makes the overall features to be sensitive to lines and object edges. Changing the presentation format to amplitude and angle will result in unsigned gradients for each given pixel. In practice, a filter can be used to convolute over the image and in each step calculate the gradient for a given pixel. Because of the unsigned gradients, the angular values are between $0^{\circ}$ to $180^{\circ}$. If nine bins of $20^{\circ}$ each are considered, the amplitude of the gradients can be represented in respected bin based on the angular value. It is worth mentioning that if the angular value is not the center of the bin, then the amplitude is going to be divided into two bins that the angle of the gradient is closest to. If the input image has more than one channel such as RGB, then channel with the highest amplitude is chosen, and also the respective angle is used for histogram representation. 

Figure \ref{fig:HOG}(b) shows one of such histograms, with normalized amplitudes based on highest value of amplitude. This figure is taken from a batch of pixels, where there is a line passing the window, so the angular value of $120^{\circ}$ to $140^{\circ}$ has the most abundance.

As an example, the HOG algorithm output is depicted as an image in Fig. \ref{fig:HOG}(c) where a $64 \times 64$ cell is one gradient cell (it is enlarged to be seen by human eyes, also less computation).

\begin{figure}[t]
    \centering
        \includegraphics[width=0.40\textwidth]{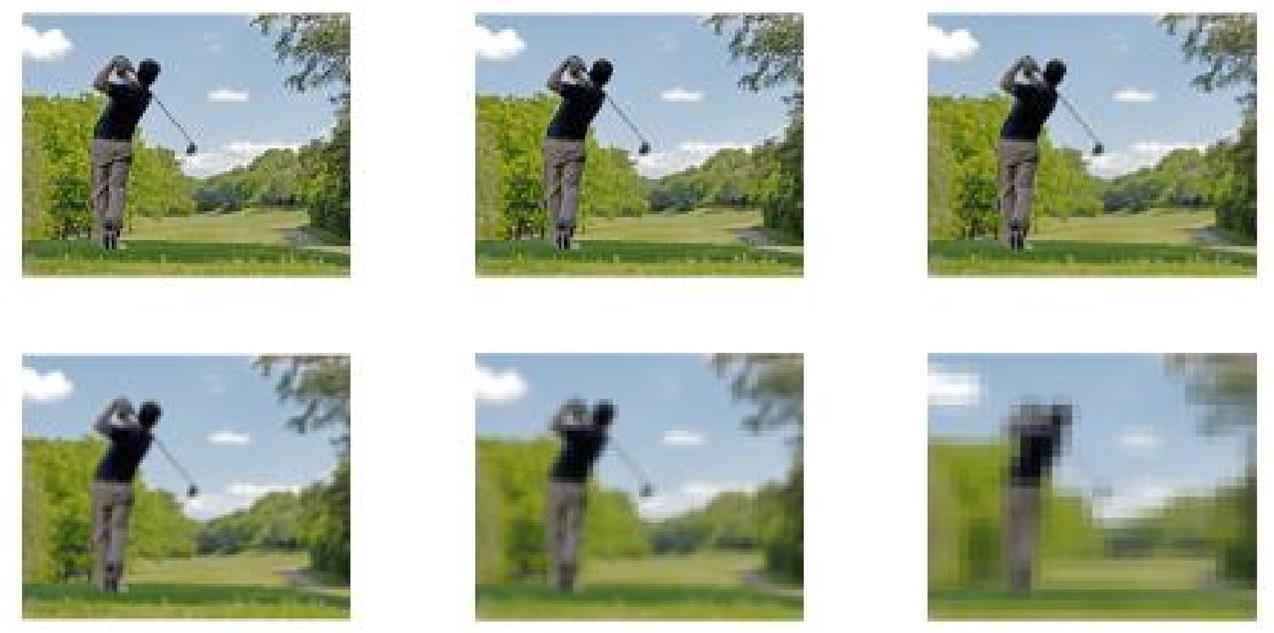}
    \caption{Image pyramid used in HOG feature extraction method.}
    \label{fig:pyramid}
    \vspace{-15pt}
\end{figure}

In an attempt to capture all details with different distances from camera location, usually a pyramid of the image is employed. The image with original resolution is considered first, and then some pixels in each row and column are discarded to create a lower resolution version of the same image, and then the same HOG algorithm will generate another feature map. The steps iterate until it is not feasible anymore to conduct classification on the image. Figure \ref{fig:pyramid} shows an image pyramid, where the top left is the actual input and the bottom-right one has the least number of pixels, but the size of each of pixels is the largest, which preserves the dimensionality of the input image.

\begin{figure}[b]
    \centering
        \includegraphics[width=0.25\textwidth]{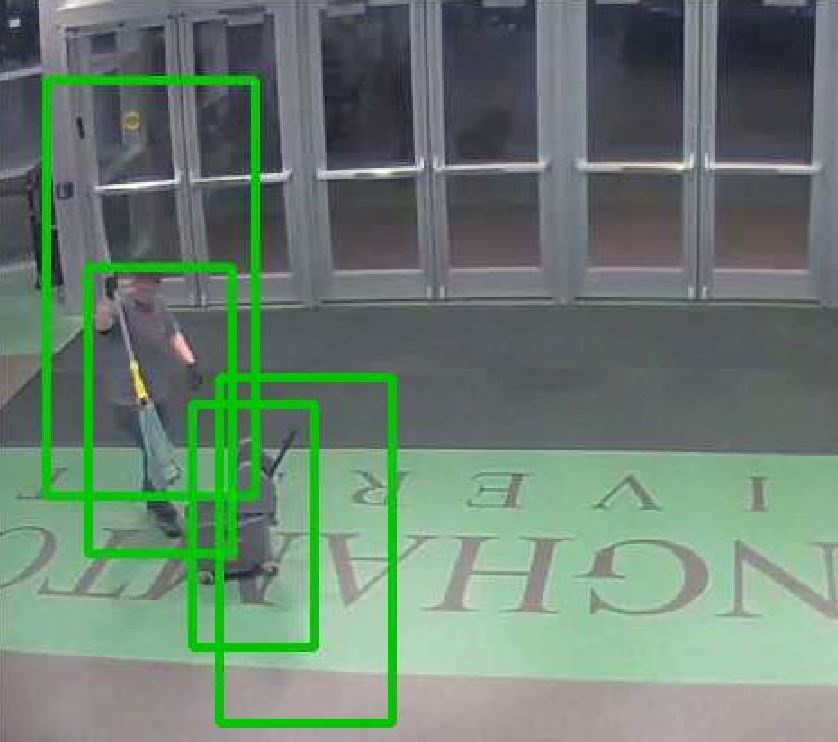}
    \caption{False multiple detection for a single human object.}
    \label{fig:hallway}
    \vspace{-10pt}
\end{figure}

The SVM classifies objects of interest at each stage, so multiple detection reports are possible. In different scenes, fine-tuning of the HOG variables might be needed to determine the number of maps generated. Figure \ref{fig:hallway} is an example, where the output detects a human object several times because several feature maps are provided to the SVM. Assuming to use the general pre-tuned variables yields an extra step to take only one of the bounding boxes and discard the rest. One mostly used method is to capture the biggest bounding box as the object. This approach may lead to an inaccurate detection. The effect is more noticeable when there are multiple human objects closer to each other. Although the detection rate can be improved by fine-tuning the filter size and variables, in practice, it is non-trivial to reconfigure once the cameras have already been installed.

As explained above, the HOG algorithm extracts features and a trained SVM based on the featues, classifies the humans. COCO image set \cite{lin2014microsoft} archive for person is used for training with around 20K images. Unfortunately, while the feature extraction presents useful information, SVM and HOG are too expensive to edge devices where these computing intensive tasks are repeatedly executed for each frame.

\section{Lightweight CNN}
\label{sec:L-CNN}

Recently, CNNs has been widely applied as a powerful tool for object classifications. However, it is considered as a challenging task to fit the CNNs into the network edge devices due to the very restrict constraints on resources. Even if the time consuming and computing intensive training can be outsourced to the cloud and the network layer architecture get simplified, edge devices still cannot afford the storage space for parameters and weight values of filters of these deep neural networks and the computation required. Therefore, a lightweight designed CNN is expected in the edge environment.

In designing the L-CNN architecture Depthwise Separable Convolution~\cite{howard2017mobilenets},~\cite{sifrerigid} is employed to reduce the computational cost of the CNN itself, without much sacrificing the accuracy of the whole network. Also, the network is specialized for human detection to reduce the unnecessary huge filter numbers in each layer. This yields to a network implementable at the edge.

\subsection{Depthwise Separable Convolution}
\label{sec:depth}

By splitting each conventional convolution layer into two parts, computational complexity is more suitable for edge devices using depthwise separable convolution and pointwise separable convolution. More specifically, the conventional convolution will take an input such as $F$, which has a dimensionality of $D_{f}\times D_{f}$ and of $M$ channels, and maps it into $G$, which is $N$ channels of $D_{g} \times D_{g}$ dimension. This is done by filter $K$, which is a set of $N$ filters, each of them is $D_{k} \times D_{k}$ and has $M$ channels, as calculated in (Eq. \ref{eq:conv}): 

\begin{equation}
 G_{k,l,n}=\sum_{i,j,m}K_{i,j,m,n} \cdot F_{k+i-1,l+j-1,m}
 \label{eq:conv}
\end{equation}

The computational complexity is

\begin{equation}
 CC_{Conventional}=D_{k} \times D_{k} \times M \times N \times D_{f} \times D_{f}
 \label{eq:comp1}
\end{equation}

\begin{figure}[t]
    \centering
        \includegraphics[width=0.2\textwidth]{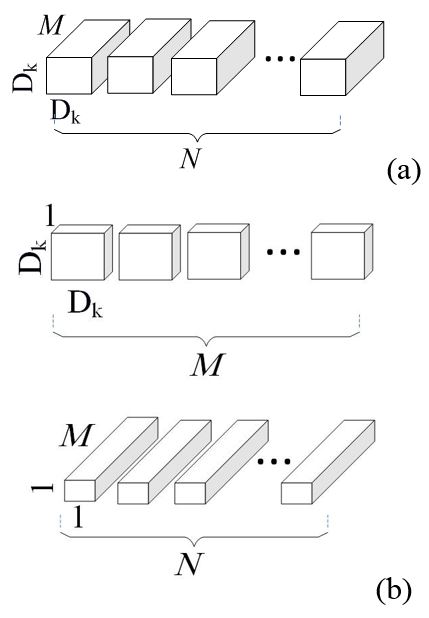}
    \caption{Comparison between (a) The conventional convolution; and (b) Depthwise separable convolution.}
    \label{fig:depth}
    \vspace{-10pt}
\end{figure}

Figure \ref{fig:depth} compares the depthwise separable convolution filters and the conventional convolution, where the same results is taken into parts to make the complexity of the operation minimized. The depthwise separable convolution consists of two parts: The first is $M$ channels of $D_{k} \times D_{k} \times 1$ filters that will generate $M$ outputs, which is a depthwise convolution layer. Next is a pointwise convolution layer in which the filters are $N$ channels of $1 \times 1$ filters. Similarly, with the input of $F$ as before this layer will produce an output such as $\hat{G}$ in (Eq. \ref{eq:depth}) the same as (Eq. \ref{eq:conv}):

\begin{equation}
 \hat{G}_{k,l,n}=\sum_{i,j,m}\hat{K}_{i,j,m,n} \cdot F_{k+i-1,l+j-1,m}
 \label{eq:depth}
\end{equation}

\noindent{where $\hat{K}$ is a depthwise convolutional filter, which has a special dimension of $D_{k} \times D_{k} \times M$ and the $m_{th}$ filter in $\hat{K}$will be applied on $m_{th}$ $F$. The computational complexity of the depthwise convolution is}

\begin{equation}
 CC_{Depth}=D_{k} \times D_{k} \times M \times D_{f} \times D_{f} + N \times M \times D_{f} \times D_{f}
 \label{eq:comp2}
\end{equation}

Based on (Eq. \ref{eq:comp2}) and (Eq. \ref{eq:comp1}), the calculation complexity is reduced by a factor calculated by (Eq. \ref{eq:comp3})~\cite{howard2017mobilenets}. It makes a faster and more efficient network that is an ideal fit for edge devices.

\begin{equation}
 \frac{CC_{Depth}}{CC_{Conventional}} = \frac{1}{N} + \frac{1}{{D_{k}}^2}
 \label{eq:comp3}
\end{equation}

Immediately after each convolutional step, there is a Batch Normalization layer or normalization and an ReLU layer for nonlinearity introduction.

\subsection{The L-CNN Architecture}
\label{sec:human}

\begin{figure}[t]
    \centering
        \includegraphics[width=0.35\textwidth]{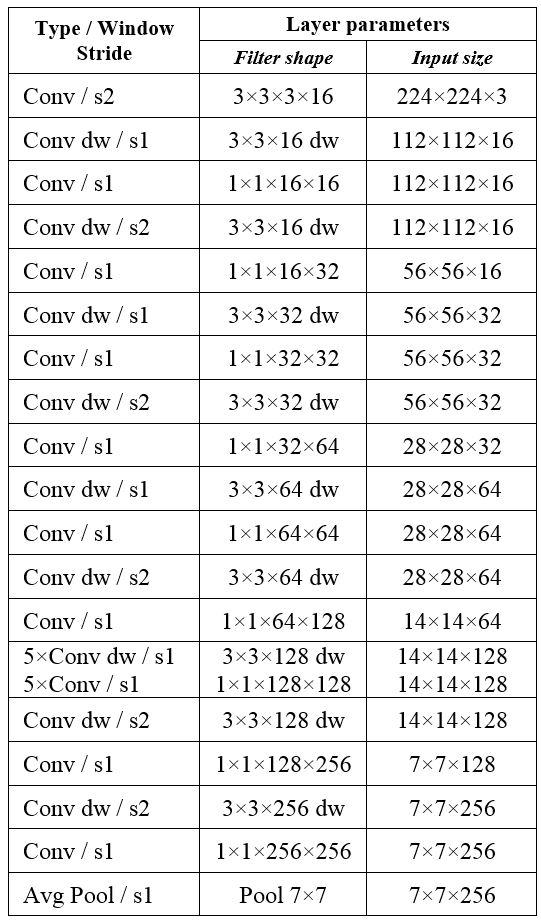}
    \caption{L-CNN network layers specification.}
    \label{fig:L-CNN}
    \vspace{-10pt}
\end{figure}

The proposed L-CNN network architecture has 23 layers considering depthwise and pointwise convolutions as separate layers, which does not count the final classifier, softmax; and regression layers to give a bounding box around the detected object. A simple fully connected neural network classifier takes the prior probabilities of each window of objects, identifies the objects within the proposed window, and adds the label for output bounding boxes at the end of the network. Figure \ref{fig:L-CNN} depicts the network filter specifications for each layer.

Downsizing happens with the help of no striding in filters and no spesific layer is added to have less computation. The first convolutional layer of the L-CNN architecture is a conventional convolution, but in the rest of the network depthwise along with pointwise convolutions are used. The L-CNN is focused on a human object detection such that the network is used only for pedestrian detection, which further simplifies the network and decreases the number of parameters to store.

Introduced in late 2016, the Single Shot Multi-Object Detector (SSD) method is faster than R-CNN~\cite{liu2016ssd} and more accurate than YOLO \cite{redmon2016you}. The name comes from the fact that in one feed forward through the network, results are generated and there is no need for extra steps taken in R-CNN. It is a unified framework for detection of an object with a single network. For training purposes, SSD architecture needs more layer architecture than conventional CNN, and when installed, it will receive the input image and output the coordination of each object detected in the image along with a label for the object. It modifies the proposal generator to get class probability instead of the existence of an object in the proposal. Instead of having the classical sliding window and checking in each window for an object to report, SSD at the beginning layer of convolutional filters will create a set of default bounding boxes over different aspect ratios, then scales them with each feature map through convolutional layers along the image itself. In the end, it will check for each object category presence, based on the prior probabilities of the objects in the bounding box and finally adjusts the bounding box to better fit the detection which means adding five layers and using outputs of two layers in SSD application. One of the downsides of SSD is that smaller objects detection accuracy is low if prior probability extraction performed in one layer. In smart surveillance, this can lead to loss of generalization. Because the goal is to detect every human object regardless of distance to camera or angle towards it. However, if the output of different feature maps from different layers is used~\cite{ren2017faster} detection rate can be increased.

%For training The images have to be the same size as the input of the network. The network accepts colored (RGB) images with the size of $224 \times 224$ pixels. Thus, for training a blob of 16 images each having three-dimensional data created and for validation blobs of the same size is used. . Lightning Memory-Mapped Database (LMDB) files were produced, which store data in a format as a \{\textit{key, value}\} pair, which leads to a faster reading speed. Furthermore, before the training, the data is normalized by calculating the mean value of each RGB channel. 

\section{L-CNN Training}
\label{sec:training}

\subsection{CNN Training}

A CNN needs to be well trained before being deployed and

applied to conduct the task of classification. Usually, the training process requires a lot of computing resources and large storage space that allows the training images be loaded and fed to the network in batches. Also the filter and other parameters should be pre-loaded into the memory. In addition, the back propagation operation incurs math intensive matrix and differential calculations. Clearly, the edge environment is not an ideal place for training. 

There are several widely used models to serve this purpose, such as TensorFlow~\cite{abadi2016tensorflow}, Keras~\cite{chollet2015keras} and Caffe~\cite{jia2014caffe}. Introduction of each gives a clear view of each platform weakness or strong points. TensorFlow is an open source software library for machine learning and artificial intelligence in general. One big benefit of this model is many GPUs that can collaborate and increase the training speed. Also, a light version of the TensorFlow is recently introduced for mobile devices, which allows loading CNN models without additional libraries needed. However, architecture in Tensorflow can be lenghty, so other platforms such as TFlearn are used to make it more compact. 

Keras gained popularity for its user-friendly, easy-to-learn environment. It uses TensorFlow as a back-end engine. Keras libraries, accessed using python, create a bridge between python syntax and TensoFlow. The libraries are created to make it easy for the user to generate and test deep modes as fast as possible. The trade-off is, allowing spontaneous coding in python, low-level flexibility of TensorFlow is sacrificed. Moreover, one of the problems that make Keras not the best choice for edge devices is that while OpenCV library supports deep learning models, it still fails to import Keras based networks. To use Keras the library itself has to be installed on the edge device, and also the results need to be loaded in the OpenCV library.

Being introduced in 2014 in C++ language, Caffe is a well-known tool for the deep learning community. It is a low-level library to work with CNNs. Fast speed makes Caffe perfect for research experiments and industry deployment. Caffe can process over 60M images per day with a single NVIDIA K40 GPU~\cite{jia2014caffe}. On the other hand, caffe has two main weaknesses. Lack of documentation on its commands makes coding a hard job. Specifically, in SSD realization of caffe model, there are layers needed for SSD deployment but very little information is provided on their functionality. Additionally, Caffe architecture is written as a plain text file, which is harder to manage when more layers are included in the architecture. 

In this work, the proposed L-CNN is trained using MXNet because of its implementation of SSD and good documentation. Apache MXNet works best for veriaty of machine learning tasks. The architecture is simplity coded through the programming language and while training, the $.jason$ file is generated. As a result the fully trained network is fast and implementable on the edge devices. LAso, MXNet is a flexible and scaleable deep learning model that many cloud centers provide today. Writing a deep learning architecture becomes an easy task because of its support of several languages such as C++, Java, Python. MXNet has a great community and documentation that will help faster reach of the results. Moreover, SSD branch of MXNet has a responsive community to help programmers learn how to familiarize themselves with SSD. The L-CNN used MXNet in python language as the structure can be generated easily in python functions. The L-CNN architecture file which is called a symbol file ready for training will have less than 80 lines with the addition of SSD code at the end. Through training phase .Jason text file of the architecture along another file containing network weights will be created~\cite{SSDMXNet} which can later be used instead of the symbol file for better speed. These outputs are closely related to what seen on caffe files (design.txt and .caffemodel).

The images from the VOC07 and VOC12 ~\cite{pascal-voc-2012} along with ImageNet~\cite{deng2009imagenet} are used to train the proposed L-CNN network where 85\% of the total set used for training and 15\% for validation. ImageNet is the biggest image set that contains more than 14 million different images from more than one thousand different classes of objects. In this application, only the images form the class of human is used. ImageNet provides bounding box coordination for some of the images in particular classes. Note that the ImageNet uses synset to name the classes, so a file with the same format of image list as VOC07 is needed. Although synset system of naming is machine-readable, it is harder for humans to understand. A combination of sub-classes for human images with coordination from ImageNet website is employed. 

The images have to be the same size as the input of the network. The network accepts colored (RGB) images with the size of $224 \times 224$ pixels. Thus, for training a blob of 16 images each having three-dimensional data created and for validation blobs of 16 images. 75\% of the total set was used for training and 25\% for validation. Lightning Memory-Mapped Database (LMDB) files were produced, which store data in a format as a \{\textit{key, value}\} pair. Converting the image set to such a format leads to a faster reading speed. Furthermore, before the training, the data is normalized by calculating the mean value of each RGB channel using Caffe platform packages. 

Training is done on a server machine with 28 CPU cores of Intel(R) Xenon(R) CPU at the base frequency of 2.4 GHz with physical memory of 256 GB. Training took 9.7 days. Several stop-criteria are introduced such as maximum iteration of 400 where each epoch is produced of 250 batches, and for every iteration, one validation test took place. also, Every 40 iterations a snapshot of the weights were created to save the progress. The training and error are calculated as Eq. \ref{eq:mse}:

\begin{equation}
 MSE=\frac{1}{n}\sum_{i=1}^{n}(\hat{y}_i - y_i)^2
 \label{eq:mse}
\end{equation}

\noindent{where $\hat{y}_i$ is the value calculated by the network, and the $y_i$ is the actual value. This Mean Square Error represents the error in object detection and linear regression used for fine-tuning the bounding box used another error. }

\section{Experimental Results}
\label{sec:exp}

\subsection{Experimental Setup}

All of the above-discussed methods are implemented on an edge device, Raspberry PI 3 Model B with ARMv7 1.2 GHz processor and 1 GB of RAM. 

Raspberry PI is a single Board Computers (SBC), which run a full operating system and have sufficient peripherals (memory, CPU, power regulation) to start execution without the addition of hardware, are targeted industrial platforms such as vending machines. The Raspberry PI Foundation made the SBC accessible to almost anyone with low cost (less than \$100) through delivering Raspberry PI product family. Given merits like commodity hardware, supporting high-level programming languages (e.g., Python) and running popular variants of Unix-based operating systems, The Raspberry Pi is an ideal platform for Edge Computing. 

\begin{figure}[b]
    \centering
        \includegraphics[width=0.5\textwidth]{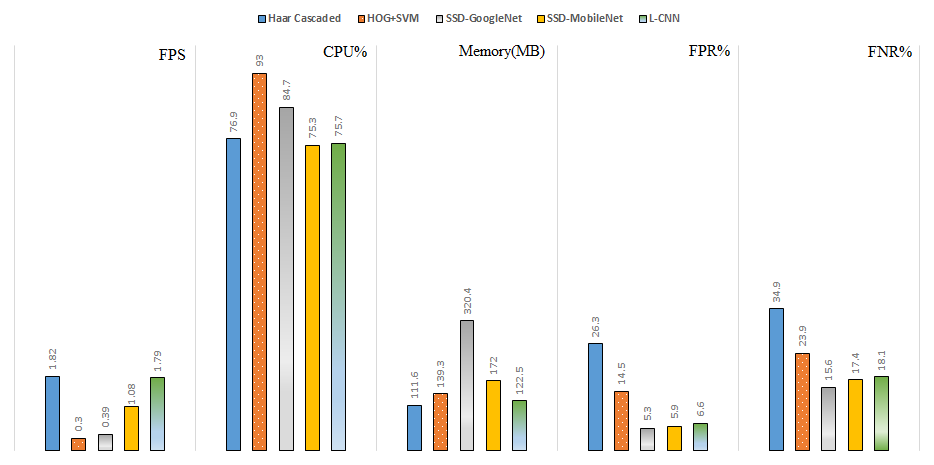}
    \caption{Performance in FPS, CPU, Memory Utility, Average False Positive Rate (FPR\%) and Average False Negative Rate (FNR\%)}
    \label{fig:percent}
    \vspace{-15pt}
\end{figure}

The CPU and memory utilization for the algorithms are captured by a third party application named memory profiler. This software is used for python applications and can track the CPU and memory used by that process. It saves the data and later plots it using python MATPLOTLIB library. Frame Per Second (FPS) is the major parameter to evaluate the performance of these algorithms. Figure \ref{fig:percent} shows the average FPS in 30 seconds of run time for each algorithm. Once again it is reminded that other CNN architectures needed to be retrained using SSD platform model so they can be used for detection rather than classification and can be compared with other object detection algorithms.

\subsection{Results and Discussions}

Figure \ref{fig:percent} summarizes the experimental results. The fastest algorithm is the Haar Cascaded, the proposed L-CNN is the second and very close to the best while other algorithms are very slow. The figure also shows that Haar Cascaded is the best in terms of resource efficiency, and again the L-CNN is the second and very close. However, in terms of average false positive rate (FPR) our L-CNN achieved a very decent performance (6.6\%) and False Negative Rate (FNR) of 18.1\% that is much better than that of Harr Cascaded (26.3\% and 34.9\%). In fact, the L-CNN's accuracy is comparable with SSD GoogleNet (5.3\% and 15.6\%), but the later features a much higher resource demanding and an extremely low speed (0.39 FPS) that makes it not suitable for edge. In contrast, the average speed of L-CNN is 1.79 FPS and a peak performance of 2.06 FPS is obtained. This is 64\% faster than MobileNet results and added along less memory usage, makes L-CNN the best choice. 

It is worth mentioning that GoogleNet does not use a huge memory portion in contrary to other reports because this is a reduced SSD based GoogleNet. As shown in Fig. \ref{fig:percent} and Table \ref{table:cnns}, with fewer classes, less parameters (thus less memory) is needed to get the same accuracy. To compute these accuracy measures, real-life surveillance video is used along with the VOC12 test dataset, and so percentages reported here may be higher than general purpose usage reported in other literature.

\begin{figure}[b]
    \centering
        \centering
        \includegraphics[width=0.4\textwidth]{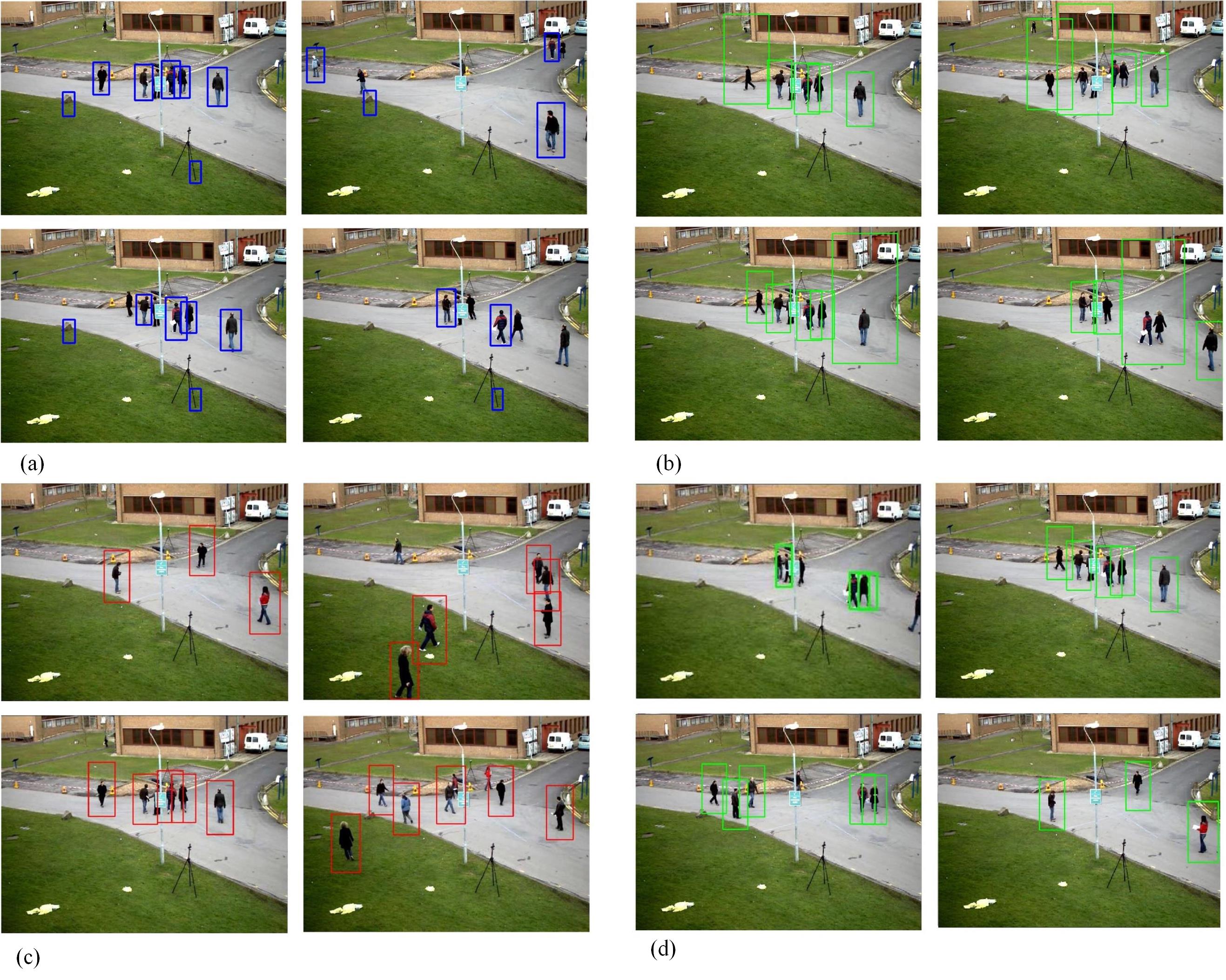}
        \caption{(a)Haar Cascaded. (b)HOG+SVM. (c)SSD-GoogleNet. (d)L-CNN.}
        \label{fig:r_haar}
\vspace{-10pt}
\end{figure}

Figures \ref{fig:r_haar} (a) to (d) show the results of Haar Cascaded, HOG+SVM, GoogleNet and L-CNN in processing a sample surveillance video. The footage is re-sized for all algorithms to $224\times224$ pixels. The smaller image size is, the less computation resource requires. Also, the deep model architecture only accepts fixed-size images. Therefore, to compare all the algorithms fairly, they all fed image with the same size. Because in practice surveillance videos are not allowed to be exposed to the public, figures included in this paper are footage from an online open source video.

The Haar Cascaded algorithm gives false detection by misidentifying the stone and the tripod as a human, shown in Fig. \ref{fig:r_haar} (a). Meanwhile, the HOG+SVM algorithm does not make the same mistakes as illustrated in Fig. \ref{fig:r_haar} (b). However, two other issues are observed. First, the bounding box is not fixed around the human objects. This may lead to inaccurate tracking performances in later steps. Secondly, in the middle of the frame where objects that are very close are considered as one in some frames, although in later frames two separate boxes are created for each person. Figures \ref{fig:r_haar} (c) and \ref{fig:r_haar} (d) verify the high accuracy achieved by the CNNs at edge.

\begin{figure}[t]
        \centering
        \includegraphics[width=0.4\textwidth]{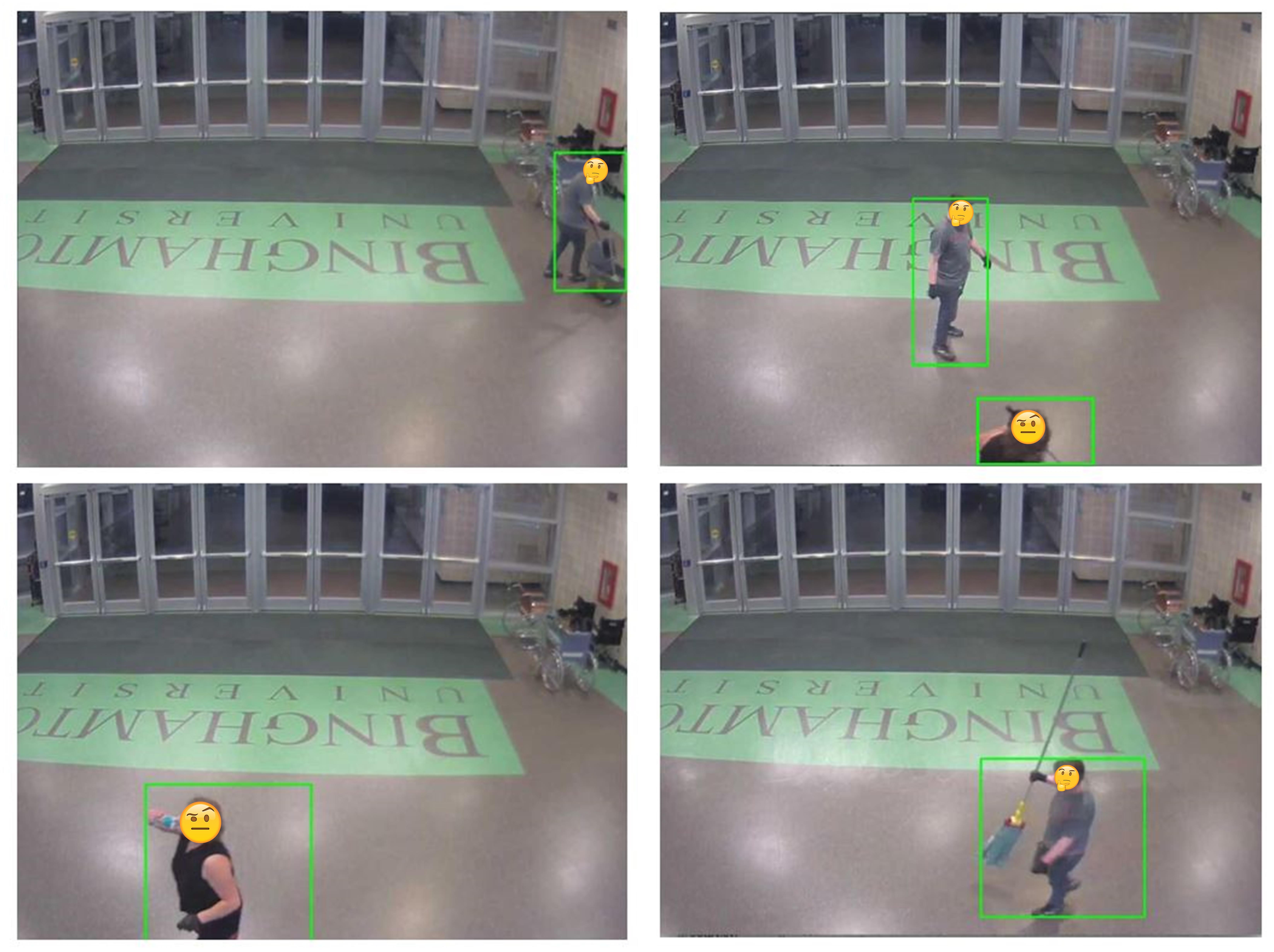}
        \caption{L-CNN: A human object from variant angles and distances.}
        \label{fig:r_L-CNN2}
\vspace{-10pt}
\end{figure}

Figure \ref{fig:r_L-CNN2} highlights the results of the L-CNN algorithm in processing video frames in which human object is captured from variant angles and distances. These are challenging scenarios for detection algorithms to decide whether or not the objects are human beings. Not only the visible features vary when the angles and distances are different, but also sometimes the human body is only partially visible or in different gestures. For example, in the right-up subfigure, the legs of the worker standing in the middle are overlapped, the second person has only head and part of the left arm captured. In the left-bottom subfigure, both two legs of the pedestrian are not visible. Many algorithms either cannot identify it is a human body, or very high false positive rate is incurred. 

Table \ref{table:cnns} compares different CNN architectures with the proposed L-CNN algorithm, including several well-known architectures such as VGG, GoogleNet, and the lightweight MobileNet. It is reminded that SSD networks because of their change in architecture and needs for special images with contour of objects for training, are dissimilar to classification CNNs and so SSD CNNs in this table are architectures that are trained using SSD. This test is performed on a desktop machine without the graphic card and with Intel(R) Core(TM) i7 (3.40 GHz and 16 GB of RAM). The result matches our intuition very well that many heavy algorithms are not good choices for an edge device as they require up to 20 times more memory space. 

\begin{table}[ht] \centering{
\begin{tabular}{|l|l|}
  \hline
  \textbf{Architecture} & \textbf{Memory (MB)} \\ \hline
  VGG & 2459.8  \\ \hline
  SSD-GoogleNet & 320.4   \\ \hline
  SqueezeNet & 145.3  \\ \hline
  MobileNet & 172.2  \\ \hline
  SSD-L-CNN & 139.5    \\ \hline
\end{tabular}
\vspace{5pt}
\caption{Memory utility of CNNs.}
\vspace{-20pt}
\label{table:cnns}
}
\end{table}

\section{Conclusions}
\label{sec:conclusions}

To make proactive urban surveillance and human behavior recognition and prediction as edge network services, timely, accurate human object detection at the edge is the essential and first step. While there are many algorithms for human detection, they are not suitable for edge computing environment. In this paper, leveraging the Depthwise Separable Convolutional network, a lightweight CNN architecture is introduced for human object detection at the edge. This model was trained using VOC07 datasets which contains the coordination of the objects of interest.  MXNet platform for neural networks was used for training, and later OpenCV libraries are used for implementation on the edge device. 

This paper has also studied the advantages and constraints of two widely used human object detection algorithms, namely Haar Cascaded object detector and HOG+SVM human detector, in the context of edge computing. Along with GoogLeNet, they are implemented on a Raspberry PI as an edge device for a comparison study. The experimental results have verified that the proposed L-CNN algorithm has met the design goals. The L-CNN has achieved satisfactory FPS (Maximum 2.03 and Average 1.79) and high accuracy (false positive rate of 6.6\% and false negative rate of 18.1\%), it uses two times fewer parameters than GoogleNet and occupies 2.6 times less memory than SSD GoogleNet. 

With the capability of immediate human object identification, our on-going efforts include the following tasks: (1) lightweight object tracking, (2) human behavior recognition, (3) suspicious activity prediction and early alarm, and (4) video clip marking for batch replay. Our ultimate goal is a proactive surveillance system that enables a more safe and secure community by identifying suspicious activities and raising alert before damages are caused.

\ifCLASSOPTIONcaptionsoff
  \newpage
\fi



%\bibliographystyle{IEEEtran}

\bibliographystyle{IEEEtranS}

\bibliography{./L-CNN}

%\begin{biography}[{\includegraphics[width=1in,height=1.25in,clip,keepaspectratio]{mshell}}]{Michael Shell}

%\begin{IEEEbiography}%[{\includegraphics[width=1in,height=1.25in,clip,keepaspectratio]{picture}}]{John Doe}
%\blindtext
%\end{IEEEbiography}


%\vfill


\end{document}